\documentclass[runningheads]{llncs}

\usepackage{amsmath}
\usepackage{graphicx}
\usepackage{siunitx}
\usepackage[pagebackref=true,colorlinks=true,breaklinks=true,bookmarks=false]{hyperref}

\emergencystretch=\maxdimen
\usepackage{lipsum}

\begin{document}

\title{Position and Rotation Invariant Sign Language Recognition from 3D Kinect Data with Recurrent Neural Networks}
\titlerunning{Sign Language Recognition}
\author{
  Prasun Roy \inst{1} \and
  Saumik Bhattacharya \inst{2} \and
  Partha Pratim Roy \inst{2} \and
  Umapada Pal \inst{1}
}
\authorrunning{P. Roy et al.}
\institute{
  CVPR Unit, Indian Statistical Institute, Kolkata, India\\
  \email{prasunroy.pr@gmail.com, umapada@isical.ac.in}\\
  \and
  Indian Institute of Technology, Roorkee, India\\
  \email{\{saumikfec, proy.fcs\}@iitr.ac.in}\\
  \url{https://github.com/prasunroy/sign-language}
}

\maketitle

\begin{abstract}
Sign language is a gesture-based symbolic communication medium among speech and hearing impaired people. It also serves as a communication bridge between non-impaired and impaired populations. Unfortunately, in most situations, a non-impaired person is not well conversant in such symbolic languages restricting the natural information flow between these two categories. Therefore, an automated translation mechanism that seamlessly translates sign language into natural language can be highly advantageous. In this paper, we attempt to perform recognition of 30 basic Indian sign gestures. Gestures are represented as temporal sequences of 3D maps (RGB + depth), each consisting of 3D coordinates of 20 body joints captured by the Kinect sensor. A recurrent neural network (RNN) is employed as the classifier. To improve the classifier's performance, we use geometric transformation for the alignment correction of depth frames. In our experiments, the model achieves 84.81\% accuracy.

\keywords{Sign language recognition \and Recurrent neural networks.}
\end{abstract}

\section{Introduction}\label{sec:introduction}

\begin{figure}[t]
    \centering
    \includegraphics[width=\linewidth]{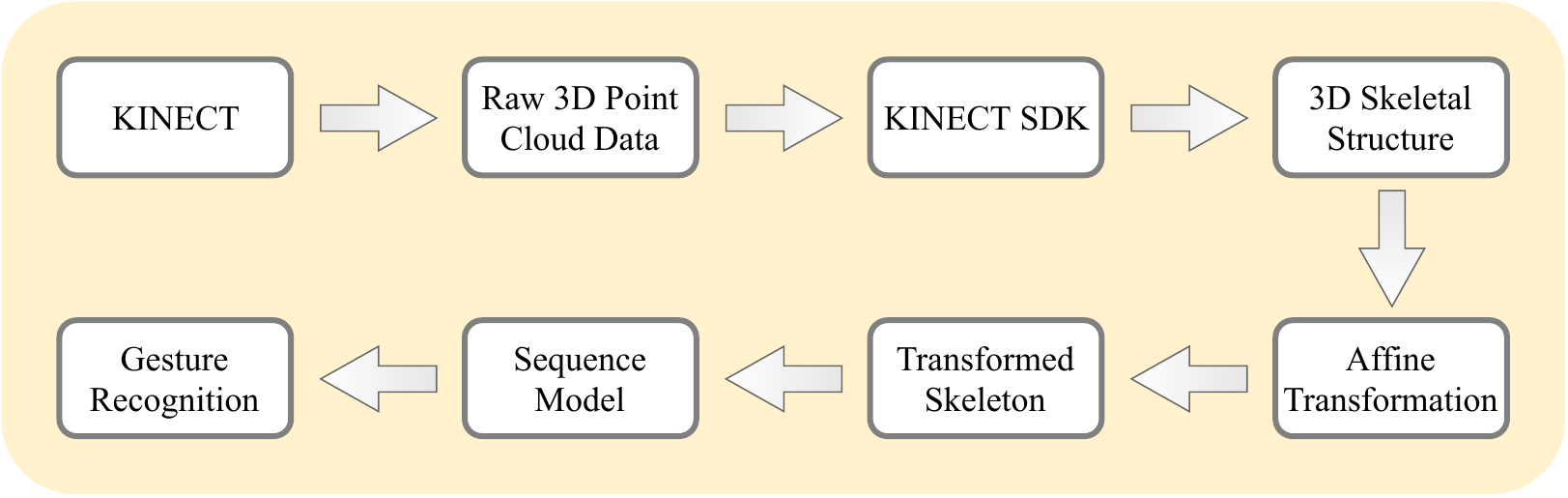}
    \caption{General workflow of Kinect-based SLR systems.}
    \label{fig:general_workflow}
\end{figure}

Sign language generally involves manual and non-manual gestures. Manual sign gestures include upper body movements, hand gestures etc., and non-manual sign gestures include facial expressions, eye movement etc. A Sign Language Recognition (SLR) system is a machine translation scheme that translates sign language into natural language and vice versa. Therefore, such systems act as a bidirectional communication channel between hearing and speech impaired and non-impaired populations. Similar to their natural language counterparts, different forms of sign language exist due to independent geographical and social contexts, such as American \cite{zafrulla2011american,chong2018american}, Arabic \cite{aliyu2016arabie}, Australian \cite{potter2013leap}, Brazilian \cite{yauri2016brazilian}, British \cite{bird2020british}, Chinese \cite{chai2013sign}, Greek \cite{ong2012sign}, Indian \cite{kumar2018position}, Spanish \cite{incertis2006hand}, Taiwanese \cite{lee2016kinect} etc.

Several attempts for building real-time SLR systems have emerged with the rapid development of cost-efficient depth sensors in recent years, such as Kinect \cite{kinect} and LEAP Motion \cite{leapmotion}. These devices include specialized sensor units and software development kits (SDK) to construct a 3D map of the environment. Recent SLR systems take advantage of such hardware devices to achieve a low-cost but reliable performance throughput. The pipeline for such SLR systems generally consists of sensor hardware and SDK for data acquisition followed by SLR software routines for analysis, feature extraction and recognition from the captured data as shown in Fig. \ref{fig:general_workflow}. In an automated end-to-end scenario, a signer needs to perform sign gestures in front of the sensor with six degrees of freedom, and the system produces final recognition results after processing the data received by the sensor. This approach suffers from self-occlusion and distorted view if the signer performs a gesture at an angle with the vertical axis on the sensor plane. In \cite{kumar2018position}, the authors have shown that affine transformations on the sensor-captured 3D coordinates can address this geometric orientation problem.

In this paper, we use data acquired by a Kinect v1 sensor to estimate 20 primary body joints of a signer. The movements of these body joints are recorded as frame-wise temporal information while performing sign gestures. A long short-term memory (LSTM) based RNN is then used as a discriminator to classify the sign gestures.

The remainder of the paper is organized as follows. Sec. \ref{sec:relatedwork} briefly reviews the significant previous works on SLR systems. Sec. \ref{sec:proposedwork} presents our proposed method. The experimental results are discussed in Sec. \ref{sec:results}. We conclude the paper by summarizing the proposed work and potential scopes in Sec. \ref{sec:conclusion}.

\section{Related Work}\label{sec:relatedwork}
The rapid development of cost-efficient depth sensors \cite{kinect,leapmotion} has significantly facilitated the emergence of hardware-accelerated SLR systems. Two widely adopted approaches use Kinect \cite{zafrulla2011american,aliyu2016arabie,yauri2016brazilian,chai2013sign,kumar2018position} or LEAP Motion \cite{chong2018american,potter2013leap,bird2020british,naglot2016real,mittal2019modified} as the primary sensor. For example, Chong et al. \cite{chong2018american} have developed an American Sign Language (ASL) based alphabet and digit recognition system using a LEAP motion sensor. The recognition process is carried out by Support Vector Machine (SVM) and deep neural networks using features extracted from hand and finger motions. Likewise, a modified LSTM-based framework \cite{mittal2019modified} is proposed for continuous Indian Sign Language (ISL). The authors have used a confidence threshold to break the continuous sign sentences into isolated sign words. Several frameworks have been proposed in the recent literature to improve SLR performance, such as multimodal systems by fusing multiple devices \cite{kumar2017coupled}, decision and feature fusion \cite{kumar2018independent,xiao2019multimodal} etc. In \cite{kumar2017coupled}, authors propose a multi-sensor SLR system using Kinect and LEAP Motion where the incoming data sequences are modeled with a coupled Hidden Markov Model (HMM). In a similar work \cite{kumar2018independent,von2008significance}, the authors fuse facial expressions with hand movement in a Bayesian framework.

\section{Proposed Work}\label{sec:proposedwork}

\begin{figure}[t]
    \centering
    \includegraphics[width=\linewidth]{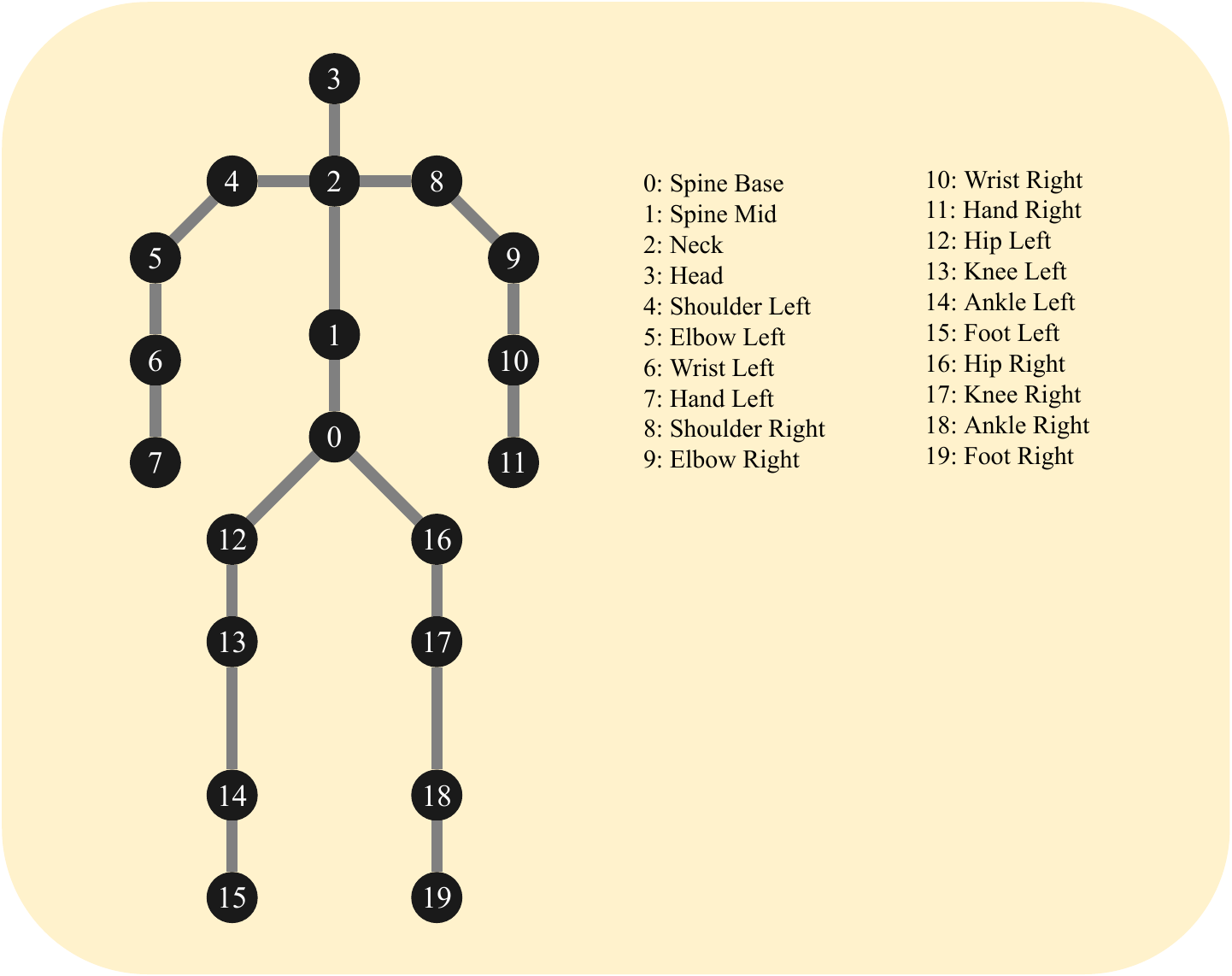}
    \caption{Enumerated body joints for Kinect v1 sensor.}
    \label{fig:body_joints}
\end{figure}

Diverse geographical and social contexts primarily influence the core attributes of any sign language. Due to such wide contextual variation, it is unlikely to have a general SLR system. This also accounts for the difficulty of comparison among these systems. Therefore, a valid comparison should involve systems for a specific sign language having a predefined vocabulary. In this study, we perform experiments on a dataset \cite{kumar2018position} of Indian sign language containing 2700 sign gestures uniformly distributed over 30 distinct categories. After acquiring depth frames using a Kinect sensor, the associated SDK constructs a skeletal structure by estimating 3D coordinates of 20 major body joints from the captured data. Afterward, the skeleton undergoes geometric transformation to minimize self-occlusion due to translation and rotation \cite{kumar2018position}. The transformed temporal data is then passed to a sequential model to assign class probabilities.

\subsection{Skeletal structure}
Kinect SDK constructs a skeletal structure of the body by estimating 3D coordinates of 20 major body joints from depth frames acquired by Kinect v1 sensor. It employs a generic camera to capture RGB frames and an IR sensor to capture depth maps. Distance $z$ of each pixel of the RGB frame from the sensor center is retrieved by querying the depth frame with spatial location $(x, y)$ of that pixel in the RGB frame. A visual representation of enumerated body joints for the Kinect v1 sensor is shown in Fig. \ref{fig:body_joints}.

\subsection{Position invariance}
A position invariant transformation helps to minimize recognition error due to the translation of 3D points in the sensor plane. We assume a signer performs gestures on $XZ$ plane and translation occurs when spine midpoint $C$ is shifted from the origin of the coordinate system $O$. Therefore, the translation vector $\overrightarrow{CO}$ is given by,

\begin{equation}
    \overrightarrow{CO} = 
    \begin{bmatrix}
        t_x\\
        t_y\\
        t_z
    \end{bmatrix} = 
    \begin{bmatrix}
        -C_x\\
        -C_y\\
        -C_z
    \end{bmatrix}
\end{equation}

Considering a point $P(x, y, z)$ in 3D, spatial coordinate $(x', y', z')$ of the translated point is given by,

\begin{equation}
    \begin{bmatrix}
        x'\\
        y'\\
        z'\\
        1
    \end{bmatrix} = 
    \begin{bmatrix}
        1 & 0 & 0 & t_x\\
        0 & 1 & 0 & t_y\\
        0 & 0 & 1 & t_z\\
        0 & 0 & 0 & 1
    \end{bmatrix}
    \times
    \begin{bmatrix}
        x\\
        y\\
        z\\
        1
    \end{bmatrix} = 
    \begin{bmatrix}
        1 & 0 & 0 & -C_x\\
        0 & 1 & 0 & -C_y\\
        0 & 0 & 1 & -C_z\\
        0 & 0 & 0 & 1
    \end{bmatrix}
    \times
    \begin{bmatrix}
        x\\
        y\\
        z\\
        1
    \end{bmatrix}
\end{equation}

\subsection{Rotation invariance}
A rotation invariant transformation is used as a heuristic to improve the performance of the SLR system by minimizing recognition error due to rotation in $XZ$ plane around $Y$ axis. This geometric transformation aims to align the body plane parallel to the Kinect sensor. We assume the spatial plane through spine midpoint $(C)$, left shoulder $(L)$ and right shoulder $(R)$ as body plane $CLR$. The unit vector $\hat{n}$ along the normal to $CLR$ plane is given by,

\begin{equation}
    \hat{n} = \dfrac{\overrightarrow{CL} \times \overrightarrow{CR}}
                    {\left\|\overrightarrow{CL} \times \overrightarrow{CR}\right\|}
\end{equation}

Angle of rotation $\theta$ is estimated as the angle between projection $\hat{p}$ of $\hat{n}$ on $XZ$ plane and unit vector $\hat{k}$ along $Z$ axis from dot product of vectors $\hat{p}$ and $\hat{k}$ as follows,

\begin{equation}
    \theta = \cos^{-1} \dfrac{\hat{p} \cdot \hat{k}}{|\hat{p}| \ |\hat{k}|}
\end{equation}

Considering a point $P(x, y, z)$ in 3D and rotation around $Y$ axis, spatial coordinate $(x', y', z')$ of the rotated point is given by,

\begin{equation}
    \begin{bmatrix}
        x'\\
        y'\\
        z'
    \end{bmatrix} = 
    \begin{bmatrix}
        \cos \theta & \quad 0 \quad & \sin \theta\\
        0 & \quad 1 \quad & 0\\
        -\sin \theta & \quad 0 \quad & \cos \theta
    \end{bmatrix}
    \times
    \begin{bmatrix}
        x\\
        y\\
        z
    \end{bmatrix}
\end{equation}

\subsection{Sequence modelling}
\begin{figure}[ht]
    \centering
    \includegraphics[width=\linewidth]{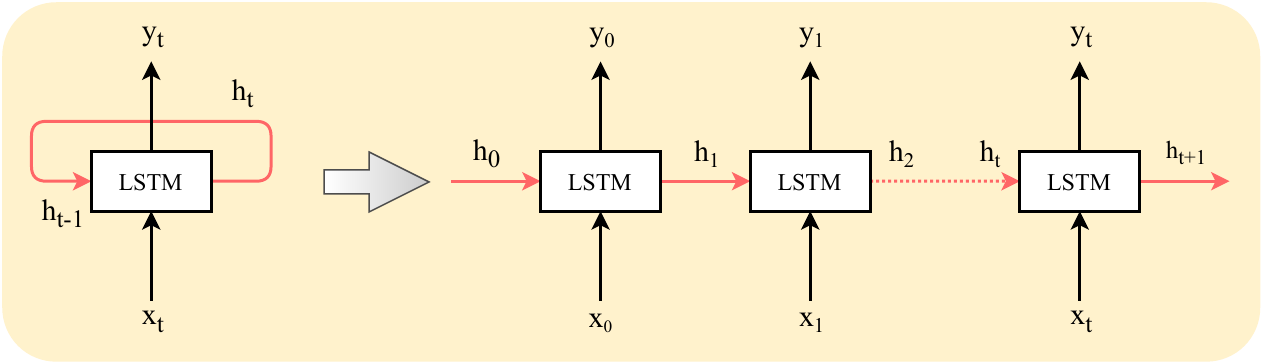}
    \caption{Architecture of a generic LSTM network.}
    \label{fig:lstm_1}
\end{figure}

Recurrent Neural Networks (RNN) are instrumental in modeling temporal data. However, they suffer from vanishing and exploding gradient problems, which restrict them from learning long-term dependencies with gradient descent \cite{bengio1994learning}. Long Short Term Memory (LSTM) networks \cite{hochreiter1997long} address these issues by introducing a few modifications in generic RNN architecture. The architecture of a generic LSTM network is shown in Fig. \ref{fig:lstm_1} where $x_i$, $y_i$ and $h_i$ denote the input, output and hidden state vectors respectively at $i$-th time step.

\begin{figure}[ht]
    \centering
    \includegraphics[width=\linewidth]{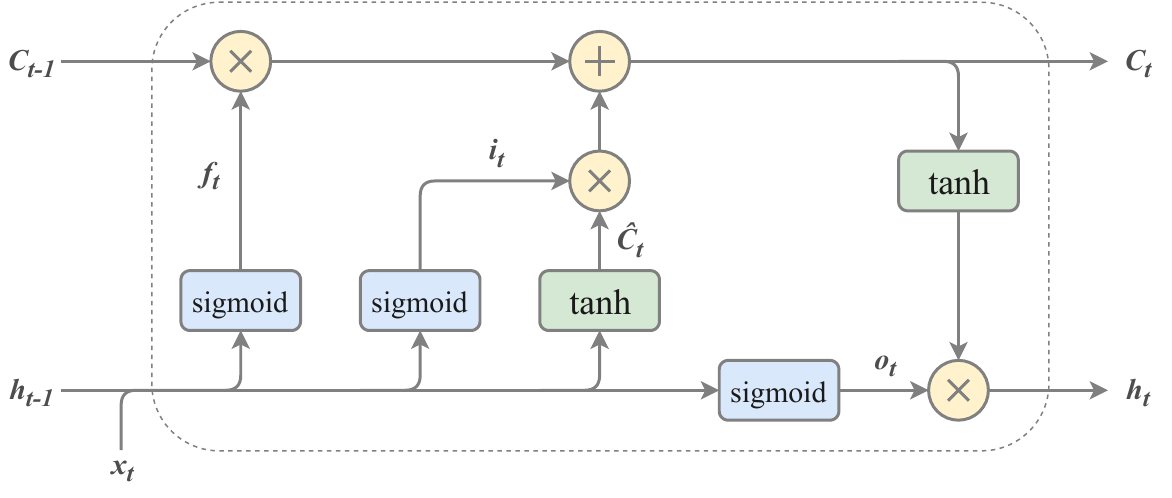}
    \caption{Architecture of a generic LSTM cell.}
    \label{fig:lstm_2}
\end{figure}

The general architecture of individual LSTM cells is shown in Fig. \ref{fig:lstm_2}. An LSTM cell consists of the forget gate, memory, and selection layers. These constituent layers of an LSTM cell are constructed using point-wise operations (addition and multiplication) and vector operations (concatenation and copy) along with activation functions (\emph{sigmoid} and \emph{tanh}).

At first, input vector of the current timestamp $x_t$ and hidden state vector of the previous timestamp $h_{t-1}$ are concatenated and passed through a $sigmoid$ activation layer (forget gate layer). The resulting vector $f_t$ thus contains values in the range $[0, 1]$ and determines which values of the previous cell state $C_{t-1}$ need to be dropped.

\begin{equation}
    f_t = \sigma(W_f \cdot [h_{t-1}, x_t] + b_f)
\end{equation}

\vspace{0.5em}

Next, another \emph{sigmoid} activation layer (input gate layer) determines the values $i_t$ that need to be updated. Also, a \emph{tanh} activation layer generates new candidate values $\overline{C_t}$ that can be added to the cell state.

\begin{align}
    i_t &= \sigma(W_i \cdot [h_{t-1}, x_t] + b_i)\\
    \overline{C_t} &= \tanh{(W_C \cdot [h_{t-1}, x_t] + b_C)}
\end{align}

\vspace{0.5em}

Next, the cell state $C_{t-1}$ is updated to $C_t$ using forget gate layer output $f_t$, input gate layer output $i_t$ and candidate vector $\overline{C_t}$.

\begin{equation}
    C_t = f_t * C_{t-1} + i_t * \overline{C_t}
\end{equation}

\vspace{0.5em}

Finally, another \emph{sigmoid} activation layer selects the parts of cell state $o_t$ to be included in the final output. Also, a \emph{tanh} activation layer maps cell state $C_t$ in the range $[-1, 1]$.

\begin{align}
    o_t &= \sigma(W_o \cdot [h_{t-1}, x_t] + b_o)\\
    h_t &= o_t * \tanh{(C_t)}
\end{align}

\vspace{0.5em}

In all of the above equations (6) -- (11), $W$ and $b$ signify \textit{weight matrix} and \textit{bias vector} of the respective layer.

\section{Results}\label{sec:results}
In this section, we discuss a detailed description of the dataset used, followed by a description of the experimental protocol and test results.

\subsection{Dataset}
In this study, we perform experiments on a dataset \cite{kumar2018position} of Indian sign language containing 2700 sign gestures uniformly distributed over 30 distinct categories. The dataset contains 16 sign gestures performed with a single hand and 14 sign gestures performed with both hands. The sign gestures are performed by 10 different signers, where each sign has been recorded 9 times by every signer. This leads to a total of $30 \times 9 \times 10 = 2700$ recorded gestures. The variation in the recorded dataset is increased by performing signs at random positions in $XZ$ plane with a rotation around $Y$ axis. Discretely, each of the 30 sign gestures is performed with a random translation along with a rotation of \ang{0} (3 times), \ang{45} (3 times) and \ang{90} (3 times) by all 10 signers to constitute a dataset of 2700 samples.

\subsection{Experimental protocol and test results}
\begin{figure}[ht]
    \centering
    \includegraphics[width=\linewidth]{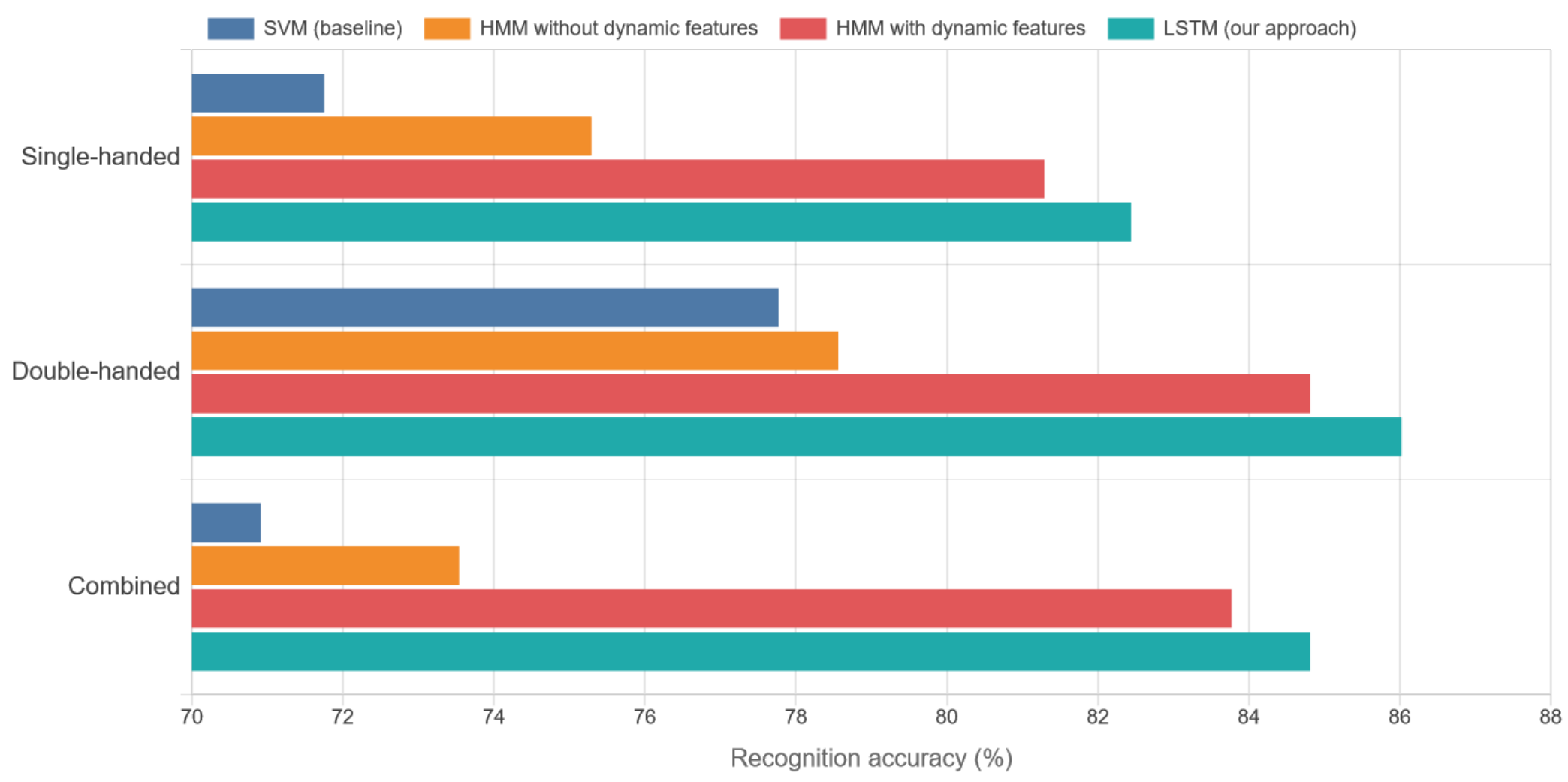}
    \caption{Quantitative comparison of recognition rates.}
    \label{fig:result}
\end{figure}

In this work, a user-independent training and validation approach is adopted where it does not require a specific signer to get registered beforehand. For a valid and direct comparison with the only known work \cite{kumar2018position} using a Hidden Markov Model (HMM) \cite{baum1966statistical} on this dataset, we perform Leave-One-Out Cross Validation (LOOCV) during training and cross-validation. In this approach, each training round uses sequentially recorded temporal data from 9 signers while keeping the data of 1 signer for cross-validation. This scheme is repeated for all 10 possible combinations, followed by taking the average of all rounds as the final validation score. In our experiments, a generic LSTM network with 100 cells shows average validation scores of 82.44\% for single-handed, 86.02\% for double-handed and 84.81\% for combined gestures. Table \ref{tab:result} and Fig. \ref{fig:result} show a quantitative comparison of recognition rates among different methods.

\begin{table}[ht]
\centering
\caption{Quantitative comparison of recognition rates.}
\label{tab:result}
\resizebox{\linewidth}{!}{
\begin{tabular}{l|c|c|c|c}
\hline
\multicolumn{1}{p{3cm}|}{\centering \textbf{\\Type of\\sign gestures\\}} &
\multicolumn{1}{p{3cm}|}{\centering \textbf{\\SVM\\(baseline)\\\cite{kumar2018position}\\}} &
\multicolumn{1}{p{3cm}|}{\centering \textbf{\\HMM without dynamic features\\\cite{kumar2018position}\\}} &
\multicolumn{1}{p{3cm}|}{\centering \textbf{\\HMM with dynamic features\\\cite{kumar2018position}\\}} &
\multicolumn{1}{p{3cm}}{\centering \textbf{\\LSTM\\(our approach)\\}} \\
\hline
Single-handed & 71.75\% & 75.29\% & 81.29\% & \textbf{82.44\%} \\
Double-handed & 77.77\% & 78.56\% & 84.81\% & \textbf{86.02\%} \\
Combined      & 70.91\% & 73.54\% & 83.77\% & \textbf{84.81\%} \\
\hline
\end{tabular}}
\end{table}

\section{Conclusion}\label{sec:conclusion}
This paper presents a position and rotation invariant, user-independent sign language recognition system using LSTM networks. The network is trained on a dataset of 2700 sign gestures uniformly distributed over 30 distinct words from the Indian sign language vocabulary and performed by 10 signers. Due to the temporal nature of the recorded sequential data, an LSTM-based recurrent neural network has been employed as the classifier. The final validation score is estimated as the average accuracy over 10-fold cross-validation. The results are compared with previous benchmarks using SVM and HMM under similar experimental protocols. We have experimentally shown that the LSTM-based approach improves the recognition rate over previous benchmarks. We believe more complex sequence models, such as bidirectional LSTM networks and attention mechanisms, can improve the current SLR performance. Additionally, a larger dataset with more variations and analyzing image frame sequences with depth frames are also worth exploring in the future.

\bibliographystyle{splncs}
\bibliography{references}

\end{document}